\documentclass[11pt,a4paper]{article}
\usepackage{authblk}
\usepackage[hyperref]{emnlp2018}
\usepackage{times}
\usepackage{latexsym}
\usepackage{graphicx}
\usepackage{url}
\usepackage{multirow}
\aclfinalcopy 
\def\aclpaperid{647} 

\title{Cross-Lingual Cross-Platform Rumor Verification Pivoting on Multimedia Content}
\author[1]{Weiming Wen}
\author[1]{Songwen Su}
\author[1]{Zhou Yu}
\affil[1]{University of California, Davis} 
\affil[ ]{\tt {wmwen@ucdavis.edu}}

\date{}
\begin{document}
\maketitle
\begin{abstract}
With the increasing popularity of smart devices, rumors with multimedia content become more and more common on social networks. The multimedia information usually makes rumors look more convincing. Therefore, finding an automatic approach to verify rumors with multimedia content is a pressing task. 
Previous rumor verification research only utilizes multimedia as input features. We propose not to use the multimedia content but to find external information in other news platforms pivoting on it. 
We introduce a new features set, cross-lingual cross-platform features that leverage the semantic similarity between the rumors and the external information. When implemented, machine learning methods utilizing such features achieved the state-of-the-art rumor verification results.
\end{abstract}

\section{Introduction}
Social network's unmoderated nature leads to the spread and emergence of information with questionable sources. With the increasing popularity of the social media, we are exposed to a plethora of rumors. Here we borrow the rumor definition from~\citet{difonzo2007rumor} as unverified information. Unmoderated rumors have not only caused financial losses to trading companies but also panic for the public~\cite{matthews2013does}. 
Especially if rumors contain multimedia content, the public generally accepts the multimedia information as a ``proof of occurrence" of the event \cite{sencar2009overview}. Readers usually don't have time to look through similar events across different platforms to make an informed judgment. Therefore, even if a credible platform, such as CNN, has debunked a rumor, it can still go viral on other social media platforms.

\begin{figure}[h]
\centering
\includegraphics[scale=0.085]{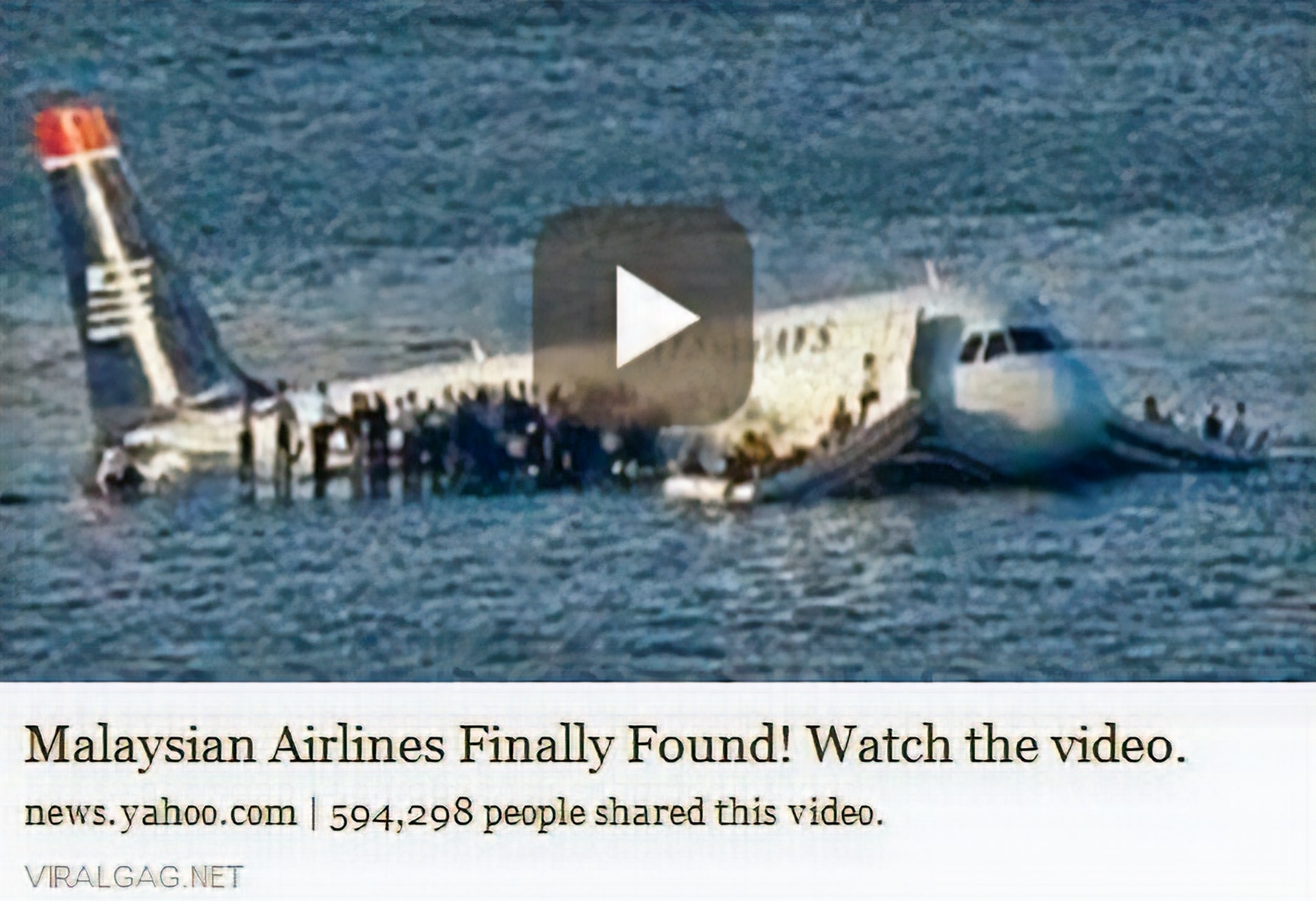}
\caption{A video of US Airways Flight 1549 was borrowed by news on Malaysia Airlines Flight 370\label{borrow}.}
\end{figure}

Intuitively, people believe fake rumors would contain fabricated multimedia content. \citet{boididou2015certh} used forensics features for detecting multimedia fabrication to verify rumors. However, these features did not lead to noticeable improvement.
We suspect that this is because un-tampered multimedia content can still convey false information when paired with fake news from a separate event.
For example, Figure \ref{borrow} shows one fake post on MH 370 that used a real video about US Airways Flight 1549.
Inspired by the fact that readers tend to search related information covered by different media outlets to garner an objective view, we propose to verify rumors pivoting on multimedia content to tackle such problems. Compared to keywords, searching information pivoting on the visual content is more effective and accurate.

In order to access information from different platforms easily, we created a new rumor verification dataset by expanding a Twitter rumor dataset to include webpages from different social media platforms using search engines. Previous rumor verification datasets are mainly monolingual, such as English~\cite{derczynski2017semeval} or Chinese~\cite{wu2015false}. However, textual information in the native language where the rumor happened can be more helpful when it comes to verifying worldwide rumors. Therefore, we not only indexed English webpages by searching Google with images but also included Chinese webpages via Baidu.

We next introduced our cross-lingual cross-platform features which capture the similarity and agreement among rumors with posts from different social media. We built an automatic verification model using the proposed features and achieved the state-of-the-art performance on the MediaEval 2015's Verifying Multimedia Use (VMU 2015) dataset~\cite{boididou2015verifying} utilizing information from Google.

Collecting and annotating rumors in foreign languages is difficult and time-consuming, especially for languages with low rumor verification labeling. Finding out an automatic way to verify those rumors in an unsupervised way is also meaningful. 
Since our cross-lingual cross-platform features are adaptable to rumors in different languages, we demonstrated that these features could transfer learned knowledge by training on one language and testing on another. Such cross-lingual adaptation ability is especially useful for predicting rumors that have low annotation resource with available annotated rumors in languages such as English.

We published our code and dataset on GitHub\footnote{https://github.com/WeimingWen/CCRV}.

\begin{table*}[h]
\centering
\small
\begin{tabular}{|l|l|l|l|l|l|l|l|l|l|}
\hline
\multirow{2}{*}{\bf ID} & \multirow{2}{*}{\bf Event}& \multicolumn{2}{|c|}{\bf Twitter} & \multicolumn{3}{|c|}{\bf Google} & \multicolumn{3}{|c|}{\bf Baidu}\\
\cline{3-10}
& &\bf Real &\bf Fake  &\bf Real &\bf Fake &\bf Others &\bf Real &\bf Fake &\bf Others\\
\hline
01&Hurricane Sandy & 4,664 & 5,558 & 1,836 & 165 & 203 & 693 & 134 & 291\\
02&Boston Marathon bombing & 344 & 189 & 619 & 54 & 49 & 317 & 55 & 16\\
03&Sochi Olympics&0 & 274 & 139 & 132& 76 & 64 & 124 & 53\\
04&MA flight 370 & 0 & 310 & 143 & 65 & 115 & 80 & 59& 31\\
05&Bring Back Our Girls & 0 & 131 & 29 & 42 & 37 & 2 & 6 &4\\
06&Columbian Chemicals& 0 & 185 & 35 & 2 & 26 & 19 & 1 &0\\
07&Passport hoax& 0 & 44 & 24 & 0 & 2 & 16 & 0 & 4\\
08&Rock Elephant& 0 & 13 & 3 & 17 & 0 & 4 & 2 & 14\\
09&Underwater bedroom& 0 & 113 & 1 & 58 & 0 & 0 & 37 & 13\\
10&Livr mobile app& 0 & 9 & 0 & 4 &11 & 0 & 0 & 0\\
11&Pig fish& 0 & 14 & 3 & 13 & 4& 1 & 12 & 7\\
12&Solar Eclipse& 140 & 137 & 40 & 64 & 39 & 0 & 10 & 91\\
13&Girl with Samurai boots& 0 & 218 & 2 & 52 & 6 & 2 & 48 & 0\\
14&Nepal Earthquake& 1,004 & 356 & 257 & 60 & 107 & 159 & 19 & 81\\
15&Garissa Attack& 73 & 6 & 60 & 0 & 3 & 36 & 1 & 0\\ 
16&Syrian boy& 0 & 1,786 & 4 & 1 & 3 & 0 & 0 & 0 \\
17&Varoufakis and zdf& 0 & 61 & 2 & 0 & 18 & 0 & 0 & 0\\\hline
&Total & 6,225 & 9,404 & 3,197 & 729 & 699 & 1,393 & 508 & 605\\\hline
\end{tabular}
\caption{\label{table:dataset}CCMR dataset statistics.}
\end{table*}

\section{Related work}\label{rw}
 Previous research has utilized multimedia information for rumor verification in various ways. \citet{zampoglou2015detecting,jin2015mcg,boididou2015certh} verified rumors by leveraging forensic features which are extracted to ensure the digital images are not tempered~\cite{sencar2009overview}. However, none of these studies found such information useful for rumor verification on a Twitter-based multimedia dataset. \citet{jin2017multimodal} incorporated image features using a pre-trained deep convolutional neural network~\cite{krizhevsky2012imagenet} on the extended Twitter-based multimedia dataset. Although the image features improve their results, their framework cannot outperform methods other than multimodal fusing networks. One possible reason is that the multimedia content in fake rumors is borrowed from another real event and usually their content corresponds to the text of the rumors. In this case, the image itself is real but not real in the context of the fake news. We thus propose to leverage the multimedia information by finding the agreement and disagreement among posts that are from different social media platforms but share similar visual contents.

The agreement between rumors and their comments is used heavily in automatic verification. \citet{mendoza2010twitter} declared that fake rumors tended to have more people question their validity. 
Later, \citet{qazvinian2011rumor} first annotated comments on tweets as supporting, denying or querying, and then used such stance information in the classification to leverage the ``wisdom of crowds". Recently, the best performing system~\cite{enayet2017niletmrg} in RumourEval shared task at SemEval 2017~\cite{derczynski2017semeval} also used such information. However, the crowd is not always wise. For example, \citet{starbird2014rumors} suspected the correctness of public opinions in rumors, pointing out some certain fake news received more support than questions. In our work, instead of using the ``wisdom of crowds", we used the knowledge from different news platforms to assist rumor verification.

Computational journalism~\cite{cohen2011computational} exploits external knowledge widely. 
\citet{diakopoulos2012finding} first leveraged information from reliable sources in the context of journalism. They developed a tool for journalists to search for and assess sources in social media around breaking news events. 
\citet{ciampaglia2015computational} utilized factual knowledge bases, such as Wikipedia, to assess the truth of simple statements.
\citet{shao2016hoaxy} designed a system for tracking rumors on different platforms, which is probably the closest work to ours. However, they did not utilize cross-platform information for rumor verification. 
Our proposed method is able to leverage information on any platform to verify rumors as long as it has both textual and multimedia information.

\begin{figure*}[h]
\centering
\includegraphics[scale=0.26]{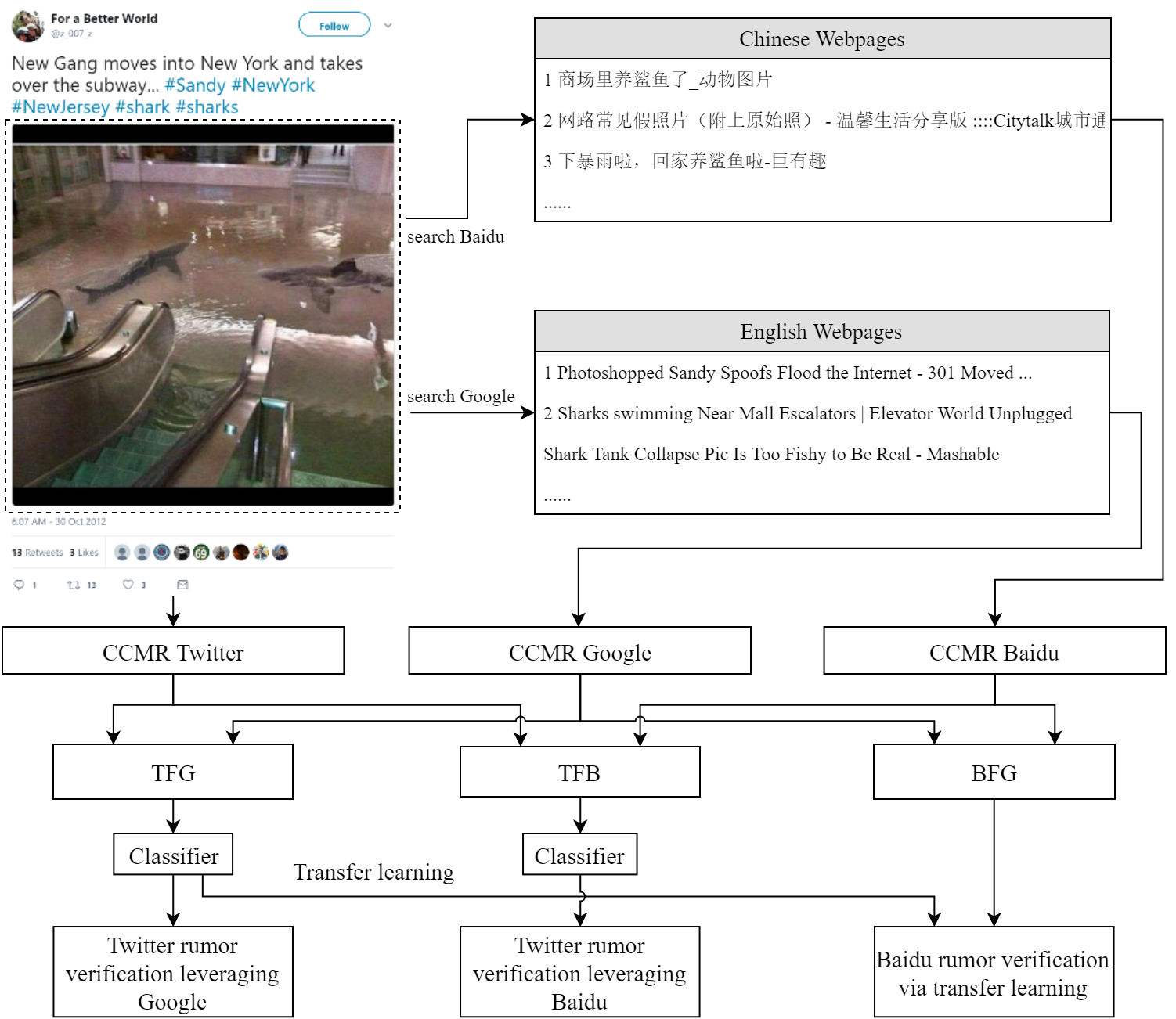}
\caption{The information flow of our proposed pipeline. TFG represents the cross-lingual cross-platform features for tweets leveraging Google information, while TFB is similar but leverages Baidu information instead. BFG means cross-lingual cross-platform features for Baidu leveraging Google information. \label{figure: structure}}
\end{figure*}

\section{CCMR dataset}\label{sec:data}

We created a cross-lingual cross-platform multimedia rumor verification dataset (CCMR) to study how to leverage information from different media platforms and different languages to verify rumor automatically. CCMR consists of three sub-datasets: CCMR Twitter, CCMR Google, and CCMR Baidu.

CCMR Twitter is borrowed from VMU 2015 dataset~\cite{boididou2015verifying}. There are 17 events containing fake and real posts with images or videos shared on Twitter. 
We created CCMR Google and CCMR Baidu by searching Google and Baidu indexed webpages that share similar multimodal content with CCMR Twitter. The upper part of Figure \ref{figure: structure} shows the collection process. We searched Google with every image (URL for video) in CCMR Twitter to get English webpages. Then we indexed those webpages to form CCMR Google. Similarly, we searched Baidu to get Chinese webpages and created CCMR Baidu.
Two human annotators manually annotated both datasets. The annotation is for better analysis and dataset quality control. None of it is utilized during our feature extraction process. Annotators were asked to label collected webpages based on their title and multimedia content. If they are not enough to tell fake news from real news, the webpage is labeled as ``others". The Cohen's kappa coefficient for two annotators is 0.8891 in CCMR Google and 0.7907 in CCMR Baidu. 
CCMR has 15,629 tweets indexed by CCMR Twitter (Twitter), 4,625 webpages indexed by CCMR Google (Google) and 2,506 webages indexed by CCMR Baidu (Baidu) related to 17 events.  
The webpages from Google and Baidu are in English and Chinese respectively. The statistics of the CCMR dataset with respect to each event is listed in Table~\ref{table:dataset}.  

\subsection{Observation in Annotation}
We observe that 15.7\% of webpages are fake in CCMR Google while 20.3\% in CCMR Baidu. We speculate that this is because all events in CCMR dataset took place outside China. Chinese webpages searched via Baidu are thus more likely to mistake the information. In the manual annotation process, we found that many images are actually borrowed from other events, which confirms our assumption. Another interesting observation is that webpages indexed by Baidu tend to have more exaggerating or misleading titles to attract click rates. 
We labeled such webpages as fake if they also convey false information through their multimedia content.

We also found that news in different languages has different distributions concerning the subtopics of the event. For example, in the Boston marathon bombing, Baidu indexed Chinese reports generally put more emphasis on a Chinese student who is one of the victims, while Google indexed English reports cover a wider range of subtopics of the event, such as the possible bomber. This phenomenon is understandable as social media from a specific country usually focus more on information related to their readers.

\section{Framework Overview}\label{sec:structure}
Figure \ref{figure: structure} describes the overview of our framework. After collecting CCMR dataset in Section \ref{sec:data}, we first performed Twitter rumor verification leveraging Google in Section \ref{sec:platform} as shown in the bottom left of the figure.
We extracted cross-lingual cross-platform features for tweets in CCMR Twitter leveraging webpages from CCMR Google (TFG). Section \ref{features} discusses the automatic construction of this feature set. We then use the features to verify rumors automatically.

We then performed Twitter rumor verification leveraging Baidu in Section \ref{sec:language} to test if our method can verify rumors by borrowing information from different languages and platforms. This experiment is meant to demonstrate that our method is language and platform agnostic. Such an advantage also enables our method to use one language information to predict in another language (see the experiment in Section \ref{sec:transfer}). We extracted cross-lingual cross-platform features for tweets leveraging webpages from CCMR Baidu instead (TFB) and used it to verify tweets in Section \ref{sec:platform}.

In Section \ref{sec:transfer}, we performed Baidu rumor verification via transfer learning to test the cross-lingual adaptation ability of the cross-lingual cross-platform features. 
We treated Chinese webpages in CCMR Baidu as rumors and empirically verified them via transfer learning. We extracted cross-lingual cross-platform features for Baidu webpages leveraging Google (BFG) in Section \ref{sec:platform}.
Since BFG and TFG are both cross-lingual cross-platform features leveraging Google,
we adopted the classifier pre-trained with TFG on CCMR Twitter to verify webpages in CCMR Baidu using BFG, under the assumption that tweets and webpages follow a similar distribution. 

Although we labeled webpages in CCMR Google and CCMR Baidu, we did not leverage the annotation here because annotation is time-consuming and using annotation information is not generalizable to other datasets.

\section{Cross-lingual Cross-platform Features}\label{features}
We propose a set of cross-lingual cross-platform features to leverage information across different social media platforms. 
We first embed both the rumor and the titles of the retrieved webpages into 300-dimension vectors with a pre-trained multilingual sentence embedding. It is trained using English-Chinese parallel news and micro-blogs in UM-Corpus \cite{tian2014corpus}. 
We encode English-Chinese parallel sentences with the same word dictionary, as they share some tokens such as URLs and punctuation. We then use a two-layer bidirectional gated recurrent unit (GRU) \cite{cho2014learning} to generate hidden states. We obtain the embedding by averaging the hidden states of the GRU. A pairwise ranking loss is used to force the cosine distance between embeddings of paired sentences to be small and unpaired sentences to be large. We train our multilingual sentence embedding on 453,000 pairs of English-Chinese parallel sentences and evaluated it on another 2000 sentence pairs. Our published code includes the implementation details of the multilingual sentence embedding.

After obtaining the embeddings of the rumor and the titles of the retrieved webpages, we further calculate the distance and agreement features between these embeddings to create a set of cross-lingual cross-platform features. In total, there are 10 features, two for distance features and eight for agreement features.
\subsection{Distance Features} 
We compute the cosine distances between the embeddings of the target rumors and the titles of the retrieved webpages. The distance indicates if the rumor has similar meaning with the retrieved webpages that have similar multimedia content. We calculate the mean and variance of the distance.

The mean of the distance indicates the average similarity between a rumor and its corresponding webpages from other platforms. A high value in mean suggests that the rumor is very different from the retrieved information from other platforms. We suspect that rumors with this property have a higher probability of being fake. Because the rumor might have borrowed the image from another event that was covered in the retrieved information. Meanwhile, the variance indicates how much these retrieved webpages are different from each other. A high variance indicates that the multimedia information is used by different events or is described in different statements. So the event or the statement the rumor covers could be fake.

\subsection{Agreement Features} 

\begin{figure}[h]
\centering
\includegraphics[scale=0.225]{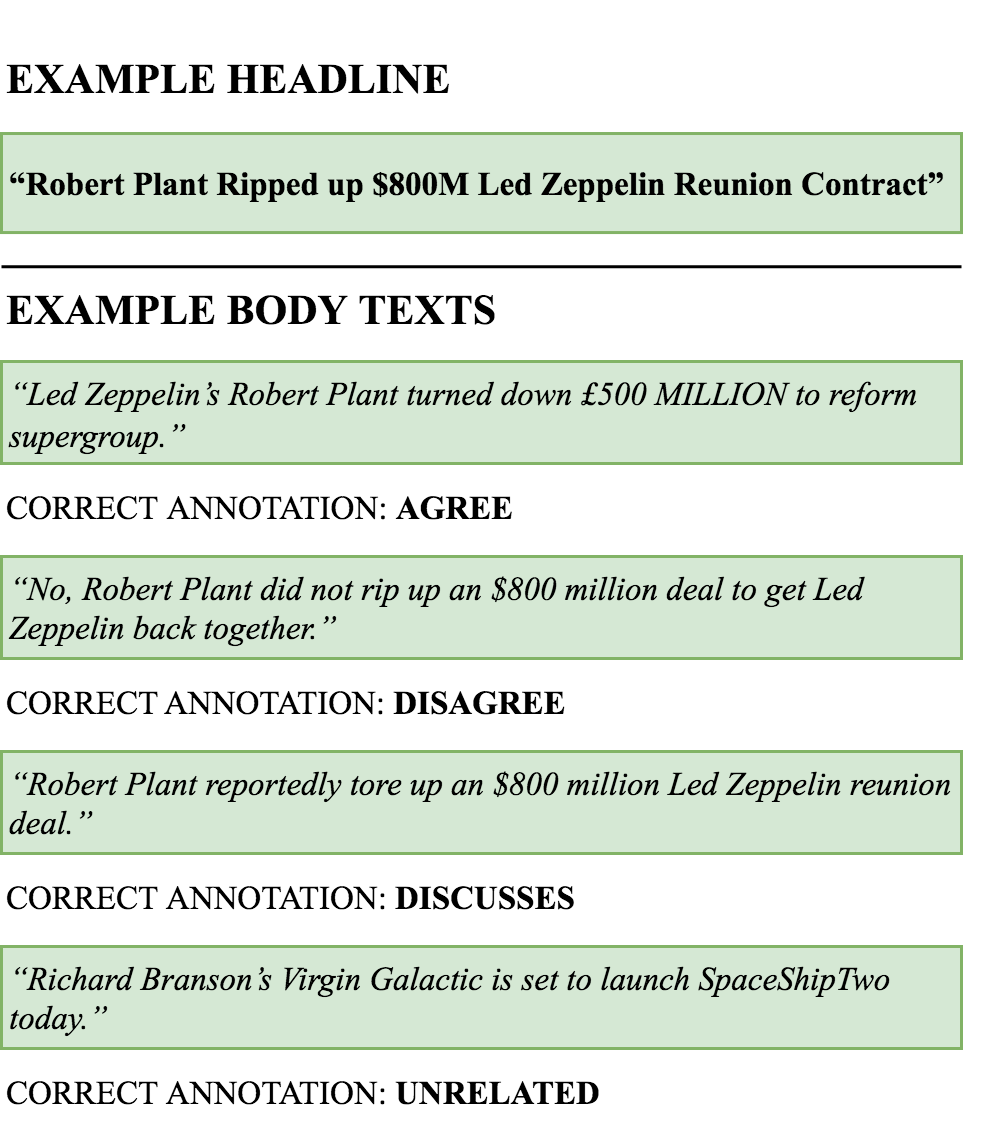}
\caption{An example headline and its body texts of the Fake News Challenge dataset.\label{figure: fnc}}
\end{figure}

We first pre-train an agreement classifier on a stance detection dataset provided by the Fake News Challenge\footnote{http://www.fakenewschallenge.org/}. This dataset provides pairs of English sentences with their agreement annotations in ``agree", ``disagree", ``discuss" and ``unrelated". Figure~\ref{figure: fnc} shows four example body texts of a headline corresponding to each type of annotation in the dataset.
During the training process, we embed the sentences in the Fake News Challenge dataset using our pre-trained multilingual sentence embedding. We then concatenate the embeddings of the sentence pair as the input. We use a multi-layer perceptron to pre-train our agreement classifier. We randomly select a balanced development set containing
250 pairs for each label (1000 in total) and train our agreement classifier on the rest of 74,385 pairs. The agreement classifier achieves 0.652 in the macro-averaged F1-score on the development set. Our published code also includes the details.

We calculate the agreement features using the mean and variance of the prediction probability between the rumor and all the retrieved webpages. There are in total four agreement labels. Therefore, we have eight agreement features in total.  
Agreement features capture information about if the rumor's statement agrees with the corresponding information in other platforms.
Besides being able to gain similar benefits as distance features, our agreement features also capture the case that the information stance is portrayed differently by different resource rumors. Conflicting information will also be an indicator of fake news.

\section{Rumor Verification Leveraging Cross-platform Information}\label{sec:platform}
We extracted cross-lingual cross-platform features for tweets in CCMR Twitter leveraging Google (TFG) and evaluated the effectiveness of TFG on rumor verification tasks.

We proposed a simple multi-layer perceptron classifier (MLP) to leverage the extracted features. MLP has two fully-connected hidden layers of 20 neurons with ReLU as the activation function. Each layer is followed by a dropout of 0.5.
We evaluated TFG in two settings: \textit{task} and \textit{event}. 1) The \textit{task} setting used event 1-11 for training and event 12-17 for testing according to \cite{boididou2015verifying}. 2) The \textit{event} setting evaluated the model performance on a leave-one-event-out cross-validation fashion. F1-score is used for evaluation metric. 
Since both collecting source rumors from Google and doing feature extraction for a given tweet can be done automatically, it is fair to compare the performance of our model with baselines described below.

\subsection{Baselines\label{baselines}}
We adopted three best performing models in the VMU 2015 task as our baselines: 

\noindent\textbf{UoS-ITI (UoS)}~\cite{middleton2015extracting} uses a natural language processing pipeline to verify tweets. It is a rule-based regular expression pattern matching method. It ranks evidence from Twitter according to the most trusted and credible sources.

\noindent\textbf{MCG-ICT (MCG)}~\cite{jin2015mcg} is an approach including two levels of classification. It treats each image or video in the dataset as a topic and uses the credibility of these topics as a new feature for the tweets. They used the tweet-based and user-based features (Base), such as the number of hashtags in the tweet or the number of friends of the user who posted the tweet.

\noindent\textbf{CERTH-UNITN (CER)}~\cite{boididou2015certh} uses an agreement-based retraining scheme. It takes advantage of its own predictions to combine two classifiers built from tweet-based and user-based features (Base). Besides features provided by the task, it included some additional features such as the number of nouns in tweets and trust scores of URLs in tweets obtained from third-party APIs.


\subsection{Results}

\begin{table}[h]
\begin{center}
\begin{tabular}{|l|l|l|l|}
\hline
\bf Method &\bf F1-Task &\bf F1-Event  \\\hline
UoS-ITI & 0.830 & 0.224 \\
MCG-ICT & \bf 0.942 & 0.756\\
CERTH-UNITN  &0.911 & 0.693 \\
TFG &0.908 & \bf0.822\\
BFG &0.810 &  0.739 \\
Combo & 0.899 & 0.816 \\
\hline
\end{tabular}
\caption{\label{table:result1} The \textit{task} and \textit{event} settings performance.}
\end{center}
\end{table}

\begin{table}[h]
\begin{center}
\begin{tabular}{|l|l|l|l|l|l|}
\hline
\bf ID & \bf UoS & \bf MCG &\bf CER &\bf TFG &\bf TFB \\\hline
01& 0.658 &0.594 &\bf 0.718  &0.715& 0.704 \\
02& 0.007 &0.494 &\bf0.745  &0.557& 0.448\\
03& 0.057 &0.882 &0.595  &\bf0.956& 0.822 \\
04& 0.538 &0.826 &0.717  &\bf0.856 & 0.678\\
05& 0.000 &\bf0.988 &0.947  &0.969 & 0.956\\
06& 0.555 &0.949 &0.916  &0.912 &\bf 1.000\\
07& 0.000 &\bf1.000 &0.475  &0.989 & 0.989 \\
08& 0.000 &0.870 &\bf1.000  &0.960 & \bf1.000 \\
09& 0.000 &0.772 &0.996  &\bf1.000 & 0.996 \\
10& 0.000 &0.615 &0.821  &\bf0.875 & - \\
11& 0.000 &\bf0.963 &0.000  &\bf0.963 & 0.667 \\
12& 0.000 &0.655 &\bf0.754  &0.677 & 0.656\\
13& 0.000 &0.954 &0.795  &\bf0.998 & 0.850 \\
14& 0.000 &0.330 &0.419  &0.409 &\bf 0.430 \\
15& 0.000 &0.130 &\bf0.156  &0.145 & 0.154 \\
16& \bf1.000 &0.990 &0.999  &0.996 & - \\
17& \bf1.000 &0.827 &\bf1.000  &0.992 & - \\ \hline    
Avg& 0.224& 0.756 &0.693 &\bf0.822  &0.739 \\ \hline
\end{tabular}
\caption{\label{table:result2} F1-scores for each event. }
\end{center}
\end{table}
We describe the \textit{task} setting results in Table \ref{table:result1}, and detailed per-event results in Table \ref{table:result2}.
Although TFG does not achieve the highest F1-score in the \textit{task} setting, it is mainly due to the split of the dataset. More than half of the tweets in the test set do not have images. Thus we can only leverage cross-platform information by searching videos' URLs, which results in less accurate cross-lingual cross-platform features.  In the \textit{event} setting, which has a more fair comparison, TFG outperformed other methods with a big margin (p\textless0.001). 
It is surprising to see that only 10 features extracted from external resources indexed by search engines leveraged by a simple classifier can bring such a big performance boost. 


To further explore the quality of the cross-lingual cross-platform features, we calculated the Pearson correlation coefficient (PCC) between each feature with respect to the tweet's label (fake or real). We evaluated both TFG and features used by baseline models.
Table \ref{table:pcc} lists the top six features with the highest absolute PCC values. A positive value indicates this feature positively correlates with fake news. We can see four out of the top six features are cross-lingual cross-platform features. The variance of the unrelated probability (unrelated variance) has the highest score, which further validates our design intuition that tweets might convey false information when they have different agreement with all other webpages that shared similar multimedia content. The second feature, ``distance var" is also highly correlated with fake news. This result supports our hypothesis that if there is a large information dissimilarity across different platforms, there is a high probability of fake information involved. The only feature from baselines (56 features in total) in the top six features is whether a tweet contains an exclamation mark or not (containsExclaimationMark). 

\begin{table}[h]
\begin{center}
\begin{tabular}{|l|l|}
\hline
\bf Feature & \bf PCC \\\hline
unrelated variance & 0.306  \\     
distance variance & 0.286\\
agree variance&  0.280 \\ 
discuss mean & -0.231 \\
unrelated mean & 0.210 \\
containsExclamationMark&  0.192  \\ 
\hline
\end{tabular}
\caption{\label{table:pcc} Top six features correlated with fake news.}
\end{center}
\end{table}

\subsection{Analysis}
We found that Google webpages usually cover a complete set of different information. There are usually both fake news and real news that debunk these fake ones. As a result, there is a big information variance among all those webpages' titles, which is captured by the cross-lingual cross-platform features. Therefore, TFG performs much better than baselines in a number of events, such as Event 03 (Sochi Olympics). The statistics of the CCMR dataset, described in Table \ref{table:dataset}, also supports our observation that the labels of posts in CCMR Google are distributed more evenly compared to other media sources. 

However, the F1-score of TFG is very low in Event 15 (Garissa Attack). Gunmen stormed the Garissa University College in this event. We analyzed the Google webpages' titles that share the same image in this event. Although some titles are related to the event, more of them are talking about completely unrelated information such as ``Daily Graphic News Sun 20th Oct, 2013 | GhHeadlines Total News ...". This webpage's title only shows its published date and the name of the website. Such noise hurt the performance of the cross-lingual cross-platform features. Since we did not perform any manual labeling or filtering, sometimes the crawled webpages can be misleading. To analyze the prevalence of such noise, we randomly picked 100 Google webpages and 100 Baidu webpages from CCMR and counted the number of noisy posts. The ratio of noise is 22\% in Google and 18\% in Baidu. However, even with such noise, our proposed methods can still outperform current state-of-the-art methods.


\section{Rumor Verification Leveraging Cross-lingual Information}\label{sec:language}
We tested if our cross-lingual cross-platform features are able to leverage external information from another language for rumor verification.
We simply replaced the Google webpages with Baidu webpages to extract features for tweets (TFB), because we have a pre-trained multilingual sentence embedding that can project Chinese and English to a shared embedding space. We used the same classifier, MLP to evaluate the performance of TFB with both baselines and TFG.

\subsection{Results}
Experiment results using the \textit{task} setting are shown in Table \ref{table:result1} and the detailed per-event results are listed in Table \ref{table:result2}. 
Similar to the problem in TFG, we can not obtain any Chinese webpages related to events such as Syrian boy, Varoufakis and zdf, which cover most tweets in the test set. Those missing features make TFB perform poorly in the \textit{task} setting.
However, TFB performs better than two of the baselines in the \textit{event} setting. If we exclude events without Baidu webpages (event 10, 16 and 17), the average F1-score of UoS, MCG and CER are 0.130, 0.732 and 0.660, which are all lower than TFB's. The performance of TFB proves that our method can be generalized across languages or platforms.

To further test the robustness of our cross-lingual cross-platform features, we also examined if it would still work when leveraging external information 
that contains different languages. We extracted the cross-lingual cross-platform features for tweets leveraging Google and Baidu webpages together (Combo) and accessed the performance of Combo using MLP similarly. The performance of Combo is also listed in Table \ref{table:result1}. Since Combo would contain noise introduced from combining webpages indexed by different search engines, it is not surprising that Combo performs slightly worse than TFG extracted from Google webpages which already cover a wide range of information solely. However, Combo performs much better than TFB which only leverages Baidu webpages. It proves that our cross-lingual cross-platform features are robust enough to utilize combined external information from different languages and platforms. 

\subsection{Analysis}
We checked the actual titles of webpages from CCMR in certain events to analyze the reason for TFB's worse performance exhaustively. We found that those Baidu webpages' titles often talk about subtopics different from the target rumor's on Twitter, while Google webpages are more related. For example, the F1-score of TFB is much lower than TFG's in event 02 (Boston Marathon bombing). This performance corresponds to our observation in Section \ref{sec:data} that Baidu webpages mainly focus on a subtopic related to the Chinese student instead of other things discussed on Twitter. 


\section{Low-resource Rumor Verification via Transfer Learning}\label{sec:transfer}
We extracted cross-lingual cross-platform features of webpages in CCMR Baidu leveraging information from Google (BFG). Then we applied Transfer, MLP in Section \ref{sec:platform} trained on the whole CCMR Twitter using TFG, to verify those webpages. Because this pre-trained model is for binary classification, only webpages labeled as real or fake in CCMR Baidu are involved.

Since webpages in CCMR Baidu do not share the same features with tweets, such as the number of likes and retweets, we adopted a random selection model as our baseline. It would predict a rumor as real or fake with the same probability. We compared the performance of the Transfer model with this baseline on each event. F1-score is also used for evaluation metric.

\subsection{Results}
Table \ref{table:transfer} lists the detailed results of our transfer learning experiment. We achieved much better performance compared to the baseline with statistical significance (p\textless0.001), which indicates that our cross-lingual cross-platform feature set can be generalized to rumors in different languages. It enables the trained classifier to leverage the information learned from one language to another.

\begin{table}[h]
\begin{center}
\small
\begin{tabular}{|l|l|l|l|}
\hline
\bf ID& \bf Event & \bf Random & \bf Transfer\\\hline 
01&Hurricane Sandy &0.247 &\bf0.287 \\
02&Boston Marathon bombing &0.230 &\bf0.284 \\
03&Sochi Olympics&0.555 &\bf0.752 \\
04&MH flight 370   &0.407 &\bf0.536\\
05&Bring Back Our Girls &0.500 &\bf0.923\\
06&Columbian Chemicals &0.000 &\bf 0.100\\
07&Passport hoax&0.000 &0.000\\
08&Rock Elephant&0.000 &\bf0.500 \\
09&Underwater bedroom&0.577 &\bf0.972 \\
10&Livr mobile app &-     &-  \\
11&Pig fish & 0.375    &\bf1.00\\
12&Solar Eclipse &0.571 &\bf0.889\\
13&Girl with Samurai boots &0.559 &\bf0.925\\
14&Nepal Earthquake &\bf0.227 &0.211  \\
15&Garissa Attack &\bf0.125 &0.059\\ 
16&Syrian boy & - &-\\
17&Varoufakis and zdf& - &-\\\hline
&Avg& 0.312 &\bf 0.531\\\hline
\end{tabular}
\caption{\label{table:transfer} Rumor verification performance on the CCMR Baidu, where - indicates there is no webpage in that event.}
\end{center}
\end{table}

\vspace {-1em}
\subsection{Analysis}
\begin{figure}[h]
\centering
\includegraphics[scale=0.5]{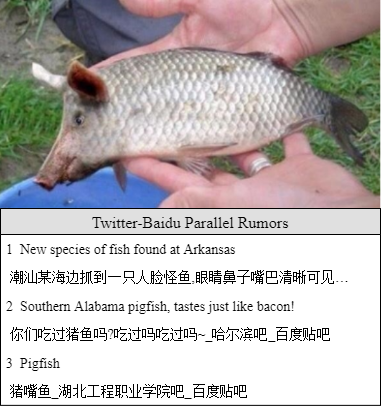} 
\caption{Example parallel rumors in the \textit{Pig fish} event\label{fig:exp3}.}
\end{figure}

In event 11 (Pig fish), Transfer achieves much higher performance than the random baseline. Generally, Baidu webpages' titles are semantically different from tweets. However, in this particular event, the textual information of those titles and tweets are semantically close. As a result, models learned from English rumors can easily work on Chinese rumors, which is helpful for our transfer learning. Figure \ref{fig:exp3} shows three Twitter-Baidu rumor pairs with similar meaning in this event. 

Transfer obtains pretty low F1-scores in event 07 (Passport hoax). The annotation conflict caused its weak performance. This event is about a Child drew all over his dad’s passport and made his dad stuck in South Korea. During the manual annotation process, we found out that it is a real event confirmed by official accounts according to one news article from Chinese social media\footnote{http://new.qq.com/cmsn/20140605/20140605002796}, while CCMR Twitter labeled such tweets as fake. Since Transfer is pre-trained using Twitter dataset, it is not surprising that Transfer achieves 0 in F1-score on this event. The annotation conflict also brings out that rumor verification will benefit from utilizing cross-lingual and cross-platform information.

\section{Conclusion}\label{conclusion}
We created a new multimedia rumor verification dataset by extending a multimedia Twitter dataset with external webpages from Google and Baidu that share similar image content. We designed a set of cross-lingual cross-platform features that leverage the similarity and agreement between information across different platforms and languages to verify rumors. The proposed features are compact and generalizable across languages. We also designed a neural network based model that utilizes the cross-lingual cross-platform features and achieved state-of-the-art results in automatic rumor verification.

\bibliography{ref}
\bibliographystyle{acl_natbib_nourl}
\newpage
\appendix


\end{document}